\documentclass[conference]{IEEEtran}
\usepackage{cite}
\usepackage{amsmath,amssymb,amsfonts}
\usepackage{algorithmic}
\usepackage{graphicx}
\usepackage{textcomp}
\usepackage{xcolor}
\def\BibTeX{{\rm B\kern-.05em{\sc i\kern-.025em b}\kern-.08em
    T\kern-.1667em\lower.7ex\hbox{E}\kern-.125emX}}

\begin{document}

\title{Robustness of Physics-Informed Neural Networks to Noise in Sensor Data \\
}

\author{\IEEEauthorblockN{Jian Cheng Wong}
\IEEEauthorblockA{\textit{Department of Fluid Dynamics} \\
\textit{Institute of High Performance Computing}\\
Singapore, Singapore \\
wongj@ihpc.a-star.edu.sg}
\and
\IEEEauthorblockN{Pao-Hsiung Chiu}
\IEEEauthorblockA{\textit{Department of Fluid Dynamics} \\
\textit{Institute of High Performance Computing}\\
Singapore, Singapore \\
chiuph@ihpc.a-star.edu.sg}
\and
\IEEEauthorblockN{Chin Chun Ooi}
\IEEEauthorblockA{\textit{Department of Fluid Dynamics} \\
\textit{Institute of High Performance Computing \&
Center for Frontier AI Research}\\
Singapore, Singapore \\
ooicc@ihpc.a-star.edu.sg}
\and
\IEEEauthorblockN{My Ha Dao}
\IEEEauthorblockA{\textit{Department of Fluid Dynamics} \\
\textit{Institute of High Performance Computing}\\
Singapore, Singapore \\
daomh@ihpc.a-star.edu.sg}
}

\maketitle

\begin{abstract}
Physics-Informed Neural Networks (PINNs) have been shown to be an effective way of incorporating physics-based domain knowledge into neural network models for many important real-world systems. They have been particularly effective as a means of inferring system information based on data, even in cases where data is scarce. Most of the current work however assumes the availability of high-quality data. In this work, we further conduct a preliminary investigation of the robustness of physics-informed neural networks to the magnitude of noise in the data. Interestingly, our experiments reveal that the inclusion of physics in the neural network is sufficient to negate the impact of noise in data originating from hypothetical low quality sensors with high signal-to-noise ratios of up to 1. The resultant predictions for this test case are seen to still match the predictive value obtained for equivalent data obtained from high-quality sensors with potentially 10x less noise. This further implies the utility of physics-informed neural network modeling for making sense of data from sensor networks in the future, especially with the advent of Industry 4.0 and the increasing trend towards ubiquitous deployment of low-cost sensors which are typically noisier.  
\end{abstract}


\section{Introduction}
Neural networks, particular in deep learning, have proven to be very successful in recent years, especially in computer vision and natural language processing tasks. However, they have been vastly less successful generally when applied in the science and engineering domain as data collection is often very difficult and a major bottleneck. 

In order to overcome this data scarcity, it has been hypothesized that incorporating knowledge about the system into these deep learning models can be a more data-efficient route for systems with known governing physics. Consequently, much work has also been published in this field in recent years \cite{stewart2017label,zhu2019physics}. 

There are many strategies for the incorporation of knowledge or physics into machine learning models, such as in the design of physics--specific model architecture or physics-guided feature selection \cite{karpatne2017theory,von2019informed,le2021surrogate}. In particular, the approach of incorporating physics-based knowledge or constraints into the model as a form of regularization has yielded very promising results across diverse domains such as fluid dynamics and electromagnetics \cite{chen2020physics,raissi2020hidden,yin2021non,chen2021physics,chiu2022can}, and is particularly effective for remedying data-absent scenarios \cite{sun2020surrogate,nabian2020physics}. More generally, the constraints embodied in these physics-based regularization can range from governing equations in the form of differential equations to simpler empirical laws such as physical equations of state.

While this use of a physics-based regularization term in the loss function in the training and construction of surrogate models for a fluid dynamical system is expected to be advantageous for data-scarce scenarios, we further explore the potential benefits of this methodology beyond improved predictive error for data-sparse or data-absent scenarios. Specifically, we describe a predominant concern over the use of purely data-driven surrogate models below, especially in the context of sparse data sets, and propose a fluid dynamical system as a case study for exploring the benefit of this methodology. 

Namely, in this current era of Industry 4.0, data acquisition frequently entails a choice between the deployment of numerous cheap, but relatively less robust and more noisy sensors at more locations, or the selective use of expensive but potentially more accurate sensors at comparatively less locations. This is exacerbated by the very specific expertise required to ensure high-quality data, for example in chemical sensing or optical imaging. This is in contrast to more intuitive domains such as computer vision, although mislabeled data can still be an issue. 

Hence, a major trade-off to be considered in the development of any data-driven machine learning model is the choice between acquisition of more noisy data, and potentially overfitting to dataset noise, or the acquisition of extremely sparse, but less noisy data. While there have been other developments in adapting the PINN framework to handle noise via Bayesian formulations \cite{yang2021b}, we choose to more directly test the ability of the physics-based regularization to compensate for both data scarcity and data noise in this work. 

In that context, we first develop a base physics-informed neural network in accordance with prior work in literature. We then evaluate the model across a series of scenarios where the data obtained and provided for training is either sparse and/or noisy, which is representative of many engineering systems where sensor systems might be expensive and difficult to deploy, and evaluate the extent to which inclusion of physics can improve the neural network’s robustness to noise in the data. 

\section{Methods}

\subsection{Fluid Dynamical Case Study}

For this work, we simulated the canonical case of a 2-D lid-driven cavity, which is a common benchmark problem in computational fluid dynamics (CFD) \cite{ghia1982high,jin2021nsfnets}. The following set of partial differential equations (PDEs) that govern incompressible steady-state fluid systems are thus considered to be physical laws that must be satisfied across the whole fluid domain $\Omega$. As these equations represent the conservation of mass and momentum, these equations are representative of physics-based laws that can be implemented in problems more generally.

\begin{subequations} 
\label{eq:ns_eqn}
\begin{align}
   &\frac{\partial u}{\partial x} + \frac{\partial v}{\partial y} = 0 \\
   &\rho\frac{\partial (uu)}{\partial x} + \rho\frac{\partial (vu)}{\partial y} = \mu\left(\frac{\partial}{\partial x}(\frac{\partial u}{\partial x}) + \frac{\partial}{\partial y}(\frac{\partial u}{\partial y})\right) - \frac{\partial p}{\partial x}  \\
   &\rho\frac{\partial (uv)}{\partial x} + \rho\frac{\partial (vv)}{\partial y} = \mu\left(\frac{\partial}{\partial x}(\frac{\partial v}{\partial x}) + \frac{\partial}{\partial y}(\frac{\partial v}{\partial y})\right) - \frac{\partial p}{\partial y}
\end{align}
\end{subequations} 

Equation 1(a) is a statement of conservation of mass, and is also called the continuity equation, while Eq. 1(b)-(c) are statements of conservation of momentum in the $x$ and $y$ direction respectively for a 2-dimensional scenario, and are collectively referred to as the momentum equations. ($u$,$v$) refers to the two components of velocity on the Cartesian grid, $p$ represents pressure, and $\rho$ and $\mu$ represent the density and viscosity of the fluid respectively. 

For this particular fluid system, the physical domain simulated is of length 1 x 1 unit. Density ($\rho$) is assumed to be of constant magnitude 1 while $\mu$ is assumed to be 0.01, hence generating a scenario with a Reynolds number of 100. A velocity Dirichlet boundary condition for the top surface was used ($u$ = 1, $v$ = 0), while the other three surfaces were specified to be stationary surfaces ($u$ = 0 and $v$ = 0). 

An incompressible CFD solver was used to solve for velocity, ($u$,$v$), and pressure \cite{chiu2018improved,chiu2021development}. The velocity and pressure fields were numerically solved on a uniform 200 × 200 grid, and subsequently down-sampled to a 50 × 50 grid to simulate a potential set of synthetic “experimental” data points that can be used for training. The original CFD-derived set of 200 × 200 grid points was used as ground truth for evaluation of test mean-squared error in all subsequent numerical experiments.

\subsection{Physics-Informed Neural Network}

A fully-connected feed-forward neural network is used for all experiments in this work. The networks for $u$, $v$, $p$ consist of 7 layers, with 32 nodes in the first layer, and 20 nodes per layer in the subsequent layers. The first four layers are shared for $u$, $v$ and $p$, while each flow variable has a unique set of parameters for the subsequent three layers. In addition, all hidden layers use the ‘sine’ activation function and a He initialization, in line with prior PINN work \cite{wong2021learning}. 

All networks are trained with ADAM, and training is terminated after $10^{5}$ iterations. An initial learning rate of 1e-3 is used which is reduced on plateauing until a minimum learning rate of 5e-6 is reached. Importantly, all data-driven neural networks and physics-informed neural networks compared in this work were trained under the same network architectures and training hyperparameters.


For the base data-driven neural network (DNN), the training loss function is defined based solely on the provided data:

\begin{subequations} 
\label{eq:dnn_loss}
\begin{align}
   &L= L_{u-data} + L_{v-data}  + L_{BC} \\
   &L_{u-data} = \frac{1}{n} \Sigma_{i=1}^{n}(u_i-u_{i,data})^2 \\
   &L_{v-data} = \frac{1}{n} \Sigma_{i=1}^{n}(v_i-v_{i,data})^2 \\
   &L_{BC}=\big\|B[u(x,y)] - g(x,y) \big\|_{\partial\Omega}^2	\\
\end{align}
\end{subequations} 

Generally, the boundary operator, $B$, can be any combination of Dirichlet or Neumann boundary conditions that enforce the desired condition $g(x,y)$ at the domain boundary $\partial\Omega$. 

The PINN training loss function includes additional physics-based regularization in contrast to the DNN training loss function above. It is defined as:

\begin{subequations} 
\label{eq:loss_fn}
\begin{align}
   &L= L_{Data} + \lambda_{PDE} L_{PDE} + L_{BC}  \\
   &L_{Data} = L_{u-data} + L_{v-data} \\
   &L_{PDE}=\big\|N_x [u(x,y)]\big\|_{\Omega}^2	\\
   &L_{BC}=\big\|B[u(x,y)] - g(x,y) \big\|_{\partial\Omega}^2
\end{align}
\end{subequations} 
which includes the data loss component, $L_{Data}$, and the BC loss components, $L_{BC}$, as per the typical data-driven neural network, and the additional physics-based regularization, $L_{PDE}$, which is determined by the set of Eqs 1(a-c). The relative weight, $\lambda$, controls the trade-off between different components in the loss function and is a hyper-parameter that can be tuned for performance optimization across different cases. A constant value of $\lambda = 1$ is used in this work.

\begin{figure}[htbp]
\begin{center}
\centerline{\includegraphics[width=0.9\linewidth]{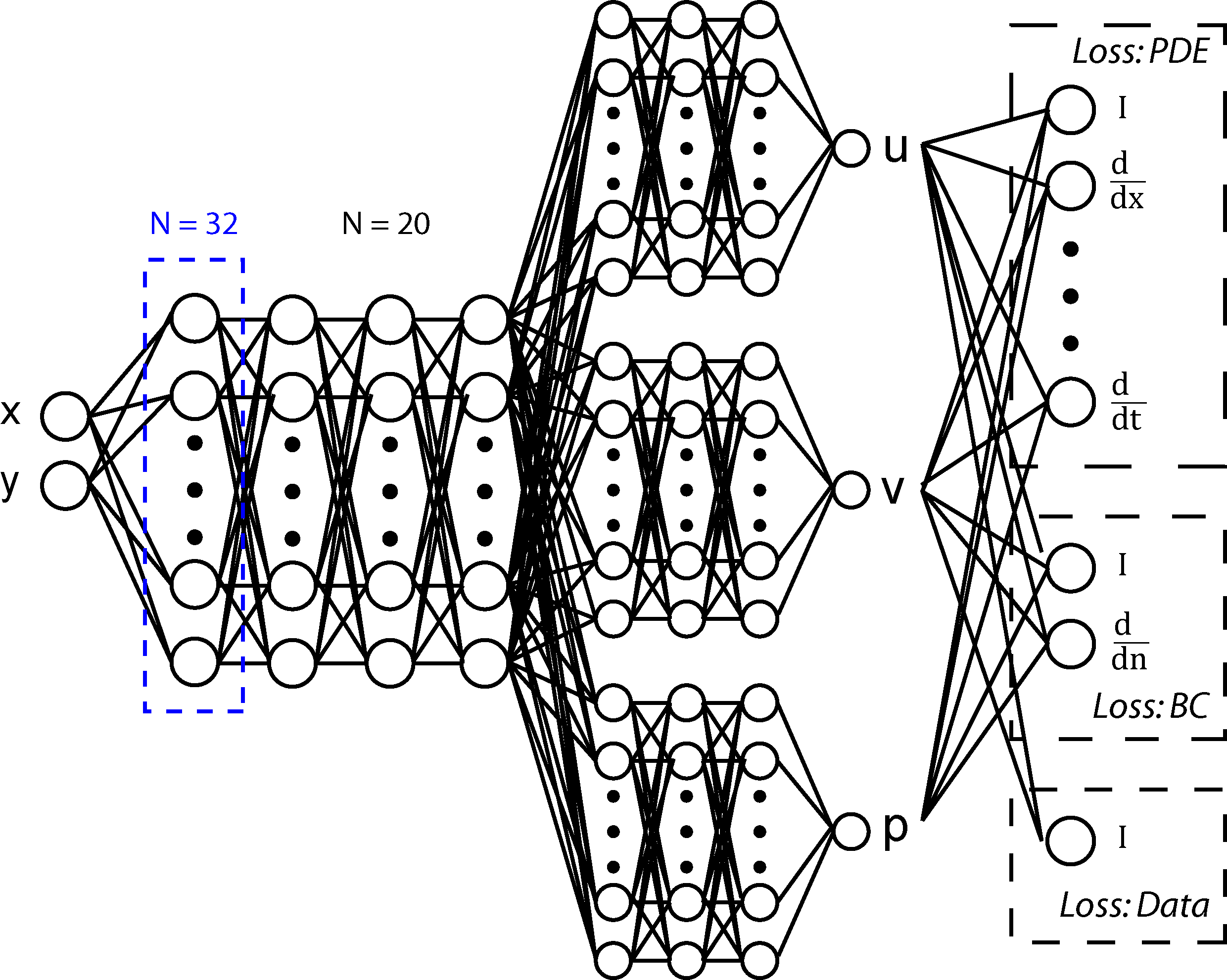}}
\caption{Illustrative schematic of the neural network architecture employed. Loss-PDE is specific only to the PINN models.}
\label{fig:nn-schematic}
\end{center}
\end{figure}

The inclusion of PDE-derived loss term, $L_{PDE}$, is typically computed by evaluating the residuals of the continuity and momentum equations in Eqs 1(a-c) via the direct use of automatic differentiation within Tensorflow. A schematic of the architecture used is presented in Fig.~\ref{fig:nn-schematic}.

In this work, the utility of incorporating physics into the neural network model is tested for situations whereby the data can be sparse and/or noisy. Hence, the training data is randomly sub-sampled from the down-sampled set of 50 × 50 points (2500 total points), and random Gaussian noise of varying amplitude is introduced to the individual points. PDE residuals are separately calculated on the 2500 locations even if no data is provided at that exact location.

Test errors are then evaluated on the original set of 200 × 200 points obtained from high-fidelity numerical simulation, which serve as the ground truth. All the different models are evaluated by the Mean-Squared Error (MSE) which is computed as:

\begin{equation} 
\label{eq:mse}
   MSE = \frac{1}{2n} \Sigma_{i=1}^{n}[(u_i-u_{i,tr})^2 + (v_i-v_{i,tr})^2]
\end{equation} 

For each experiment, the neural network’s weights are randomly initialized and trained across 10 replicates. These replicate experiments span 2 different ranges of noise and data scarcity. The distribution of MSE from model predictions across these replicate experiments are presented and discussed in the following Results section. 

\section{Results and Discussion}
While many experiments in literature have focused on demonstrating the benefits of physics-informed learning in data-absent scenarios, there has been relatively less attention devoted to an assessment of the interplay between the amount of data provided and the magnitude of noise inherent in the data. Indeed, much of the current published work focuses more on relatively noise-free data. 

\subsection{Inference with Varying Noise Magnitudes}
\begin{figure*}[htbp]
\begin{center}
\centerline{\includegraphics[width=0.8\linewidth]{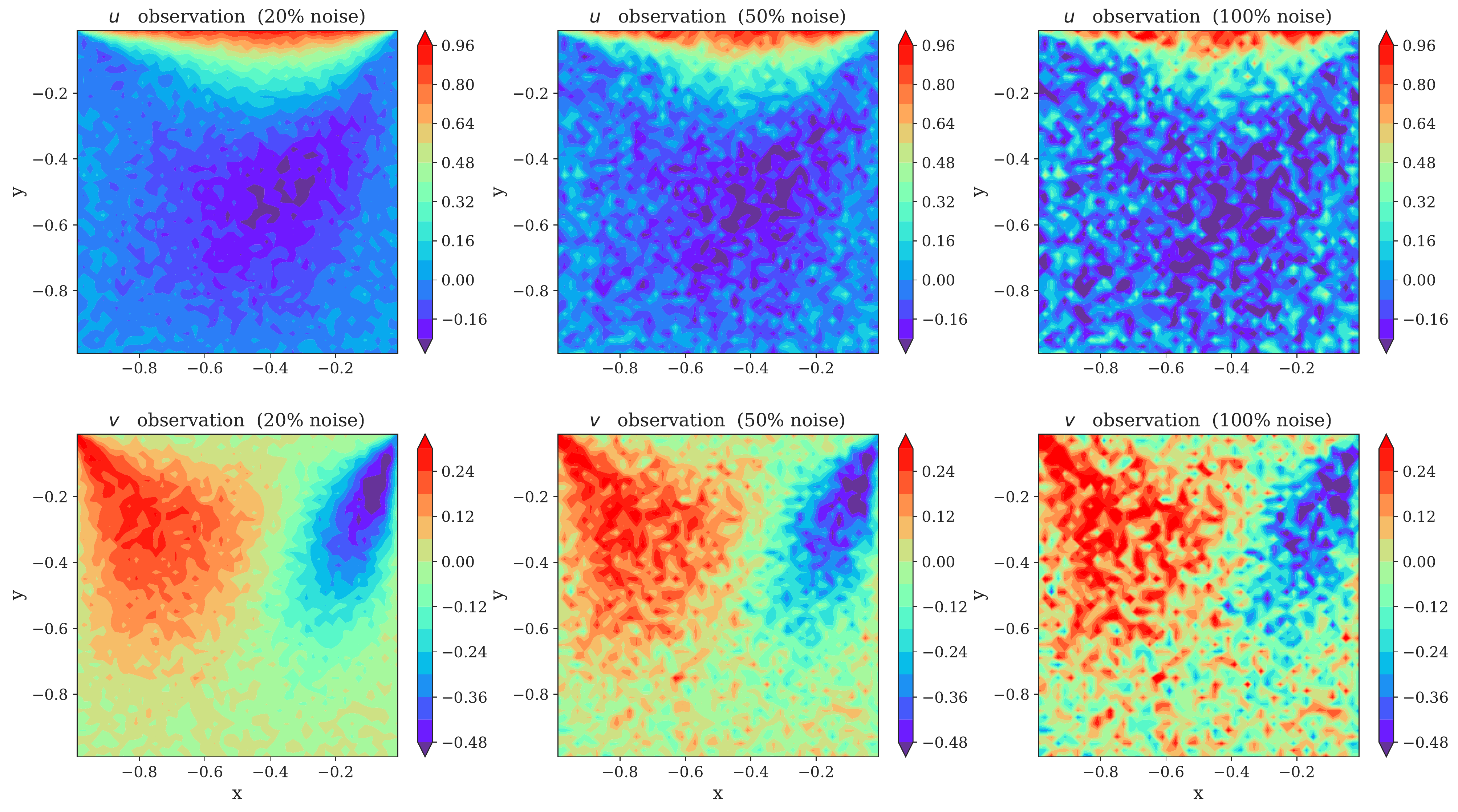}}
\caption{Contour plots of velocity (Row 1 - $u$-velocity; Row 2 - $v$-velocity) illustrating the difference between the different scenarios where (L-R) 20\%, 50\% and 100\% of noise are present in the simulated data.}
\label{fig:noise-contour-raw}
\end{center}
\end{figure*}

In the first set of experiments, noise of varying magnitudes was added to a set of 2500 data points spanning the entire fluid domain. The noise was scaled to be between 10\%, 20\%, 30\%, 40\%, 50\% and 100\% of the standard deviation inherent in the velocity data across the domain. This scaling is representative of the commonly used Signal-to-Noise Ratio (SNR) metric used to characterize the quality of sensors in many engineering applications. Importantly, it should be noted that an SNR $>$ 1 is a typical threshold beyond which any sensor is regarded to be generally unusable. Representative contour plots are presented in Fig.~\ref{fig:noise-contour-raw} to illustrate the impact of different magnitudes of noise on the data provided. In particular, it is worth noting how flow structures in the centre of the domain are completely lost due to the addition of noise.

\begin{figure}[htbp]
\begin{center}
\centerline{\includegraphics[width=0.9\linewidth]{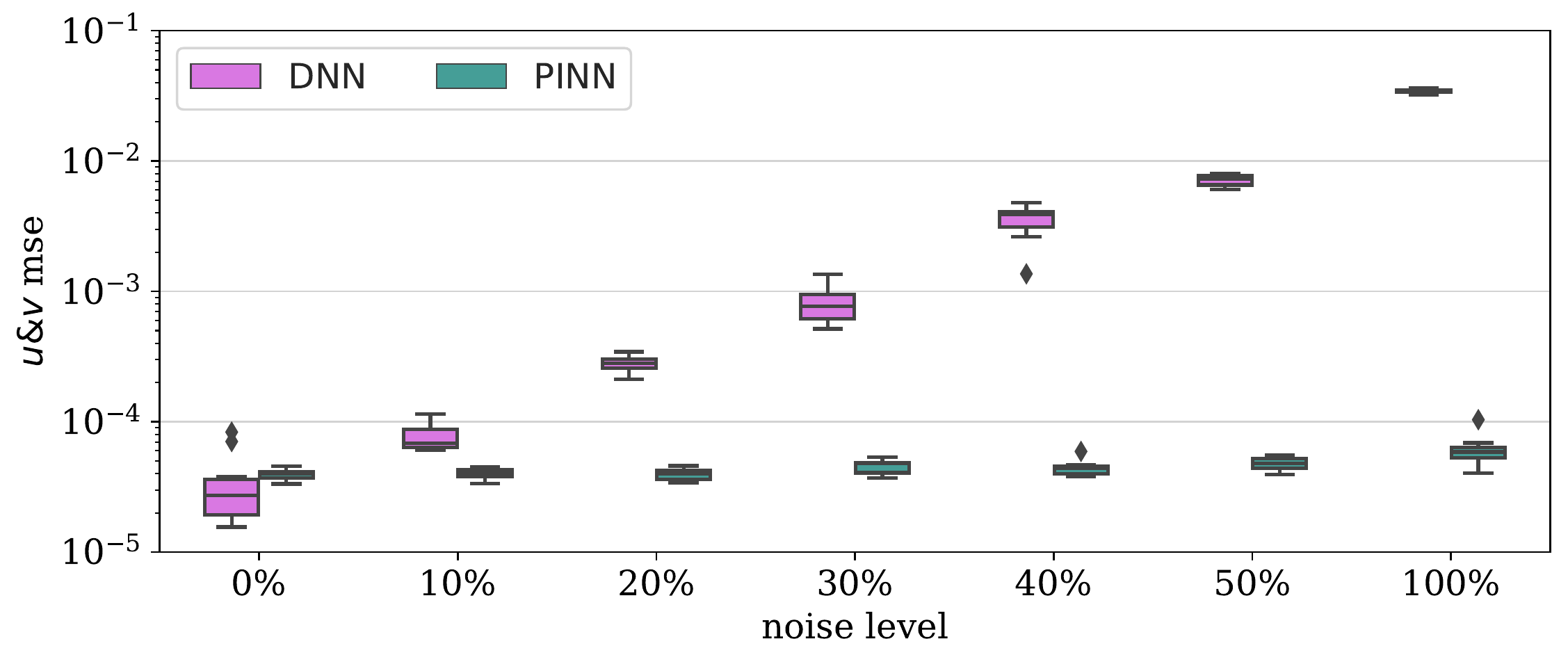}}
\caption{Boxplots of velocity test MSE for different datasets spanning sensor SNRs of varying magnitudes.}
\label{fig:noise-uv-mse}
\end{center}
\end{figure}

These data-sets representing data acquired from sensors of varying SNR are then provided to both a vanilla data-driven neural network (DNN) and a PINN. The MSEs are then computed for all the scenarios, and presented in Fig.~\ref{fig:noise-uv-mse}. 

\begin{figure}[htbp]
\begin{center}
\centerline{\includegraphics[width=0.9\linewidth]{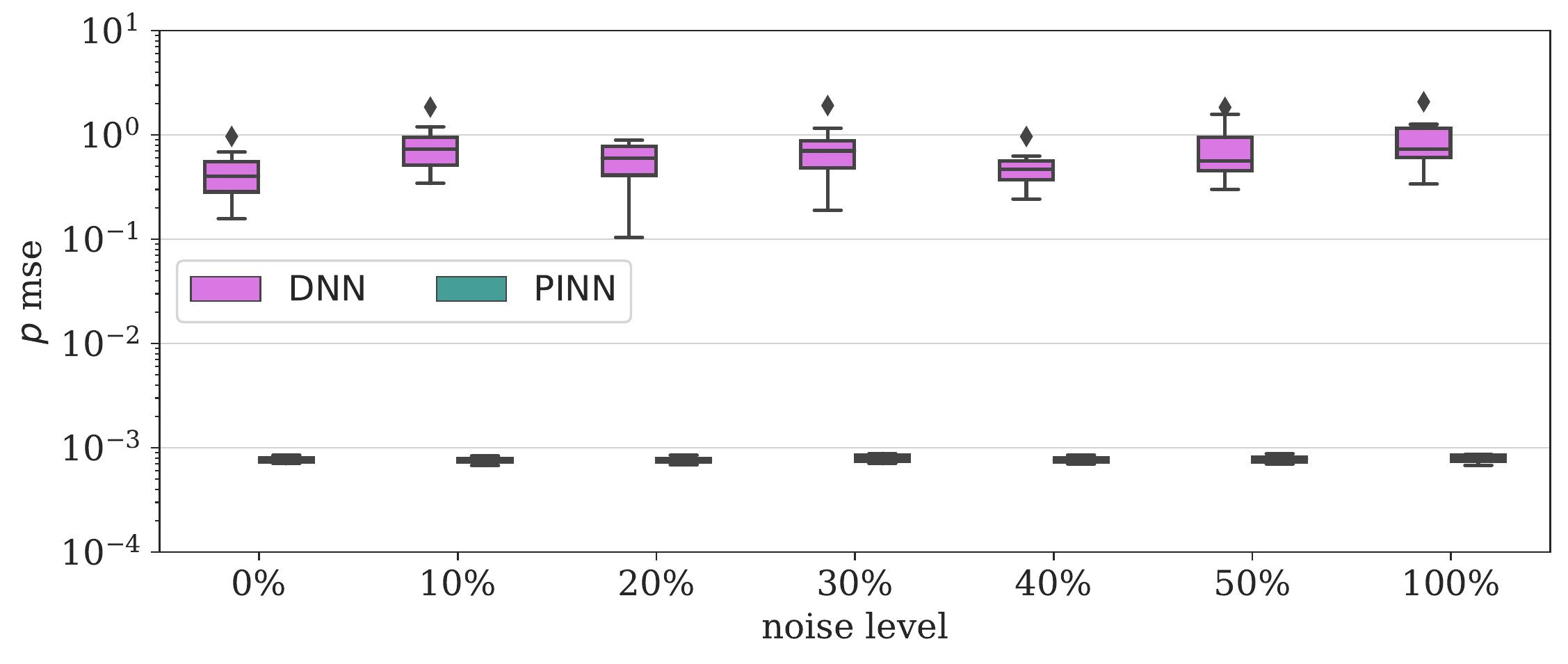}}
\caption{Boxplots of pressure test MSE for different datasets spanning sensor SNRs of varying magnitudes.}
\label{fig:noise-p-mse}
\end{center}
\end{figure}

It is also worth noting that one of the purported uses for PINNs is the ability to infer missing information about particular scenarios based on partial data. In fluid dynamics in particular, a common problem encountered in the use of particle image velocimetry for characterization of specific systems is the fact that only velocity field information can be acquired \cite{le2021u,wang2022dense}. Pressure information is typically not simultaneously acquired and must be inferred. The use of a typical DNN is not able to provide meaningful predictions for pressure unless corresponding velocity-to-pressure models are trained or corresponding numerical methods are employed. However, the use of a PINN integrates relevant underlying knowledge with the provided velocity data for highly accurate and convenient inference of pressure, as depicted in Fig.~\ref{fig:noise-p-mse}.

\begin{figure*}[htbp]
\begin{center}
\centerline{\includegraphics[width=0.9\linewidth]{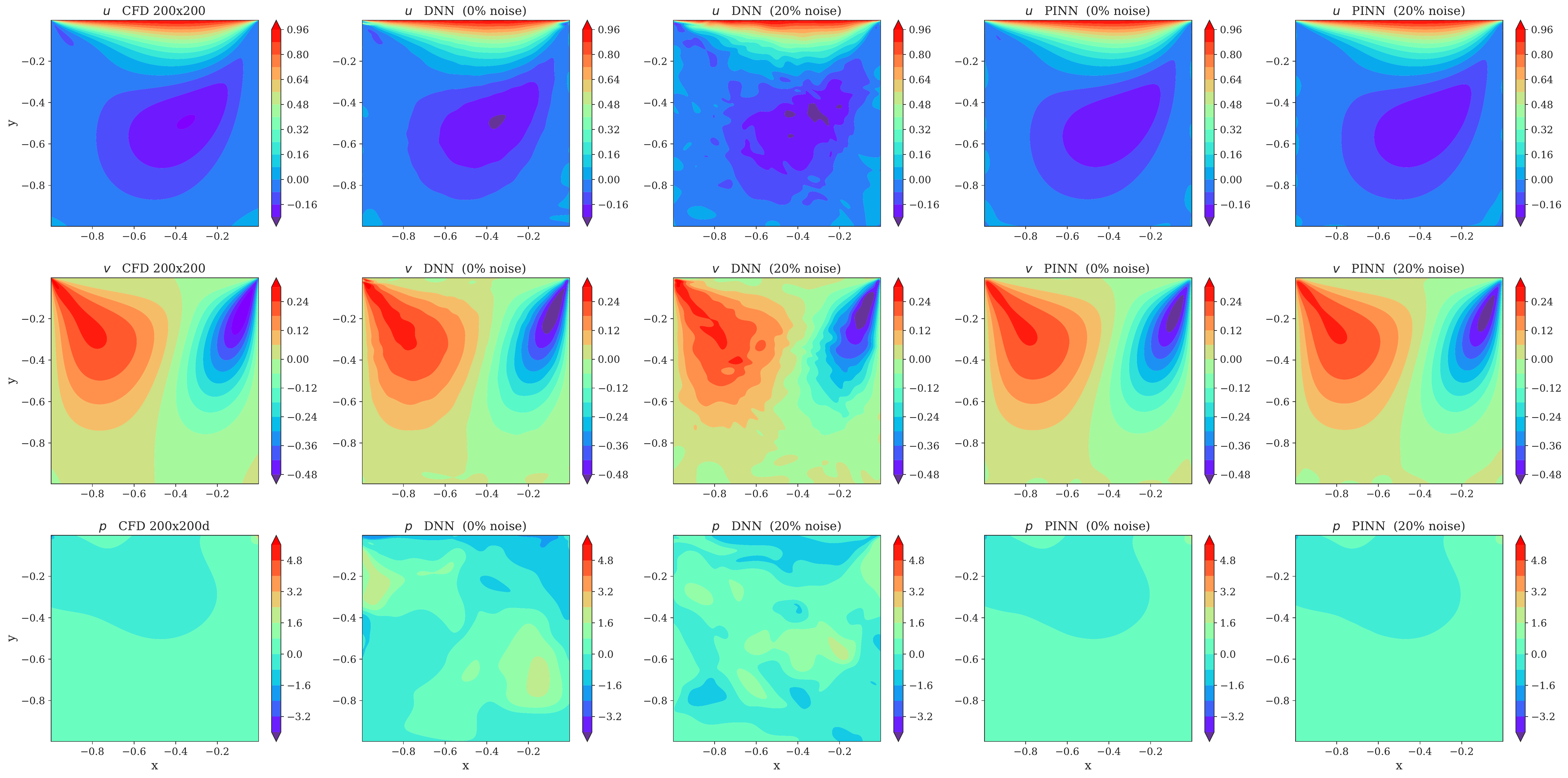}}
\caption{Representative contour plots of the $u$-velocity (Row 1), $v$-velocity (Row 2), and pressure (Row 3) for the (L-R) ground truth, DNN with 0\% noise and 20\% noise, and PINN with 0\% noise 20\% noise.}
\label{fig:contour-pinns-noise}
\end{center}
\end{figure*}

Representative contour plots of the velocity and pressure predicted by the DNN and PINN models are presented in Fig.~\ref{fig:contour-pinns-noise} to illustrate the effectiveness of the PINN models relative to the DNN models.

Unsurprisingly, an increase in SNR ratio causes a significant degradation in the predictive performance of a pure DNN. The test MSE rapidly increases by more than 2 orders of magnitude from less than $10^{-4}$ to $10^{-2}$. However, the use of a PINN model is able to compensate for the type of noise present in low quality sensors and restore them to the accuracy achievable by a typical DNN trained on data acquired from much more accurate sensors with low SNR of less than 10\%.

\subsection{Inference with Noisy Data of Variable Dataset Size}
As a further evaluation of the utility of including physics, additional experiments are conducted with down-sampled datasets of different sizes to simulate and illustrate the impact when high-quality (relatively low noise) sensors are employed but resource/experimental constraints limit the amount of data that can be acquired instead. 

\begin{figure}[htbp]
\begin{center}
\centerline{\includegraphics[width=0.9\linewidth]{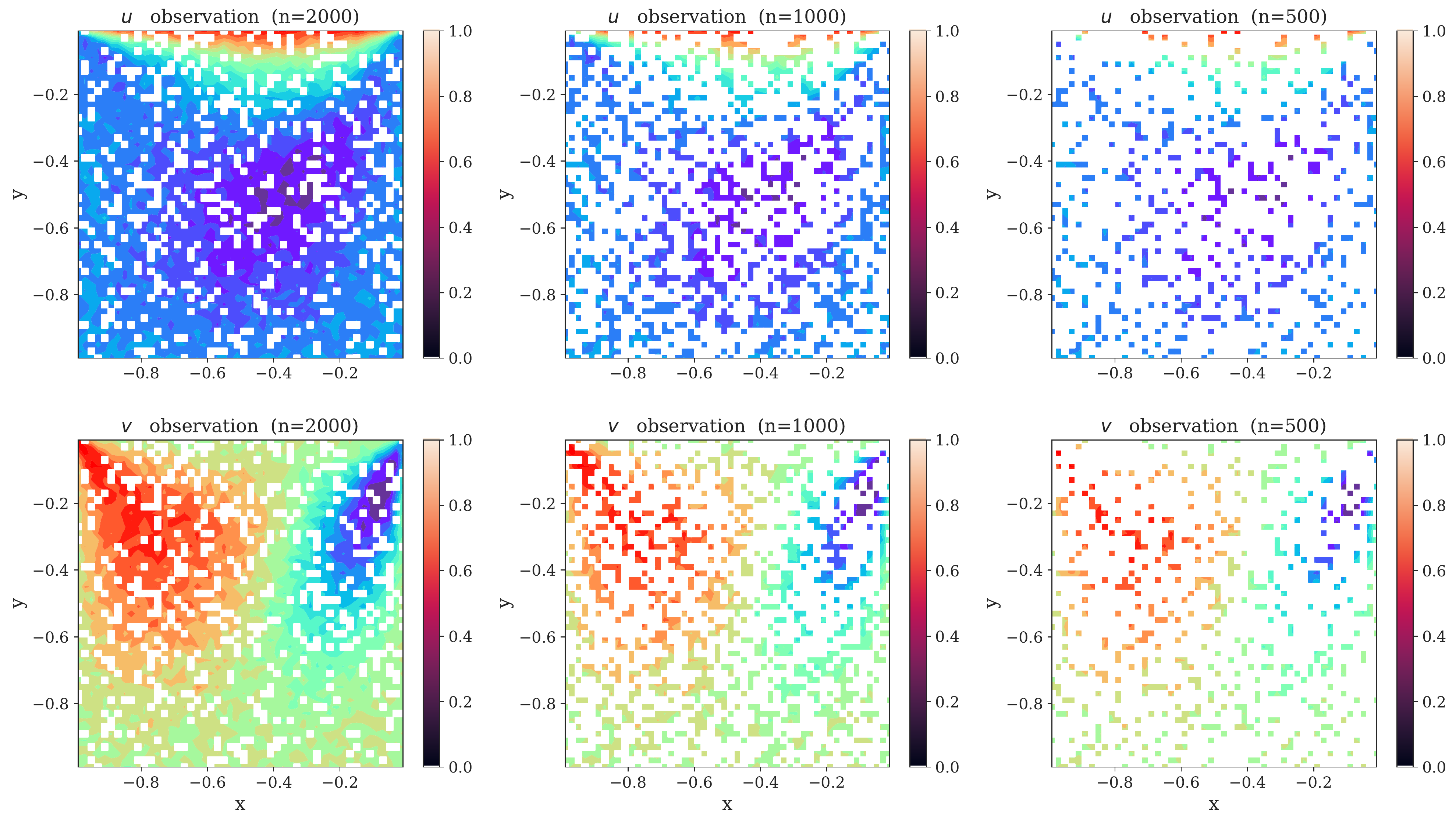}}
\caption{Contour plots of velocity (Row 1 - $u$-velocity; Row 2 - $v$-velocity) illustrating the difference between the different scenarios where 500, 1000, and 2000 data points with 10\% noise are acquired.}
\label{fig:nobs-contour-raw}
\end{center}
\end{figure}

Representative plots of the impact of smaller data-sets are presented in Fig.~\ref{fig:nobs-contour-raw}, illustrating the relative amounts of information contained when the dataset is as large as 2000 points in the domain.

\begin{figure}[htbp]
\begin{center}
\centerline{\includegraphics[width=0.9\linewidth]{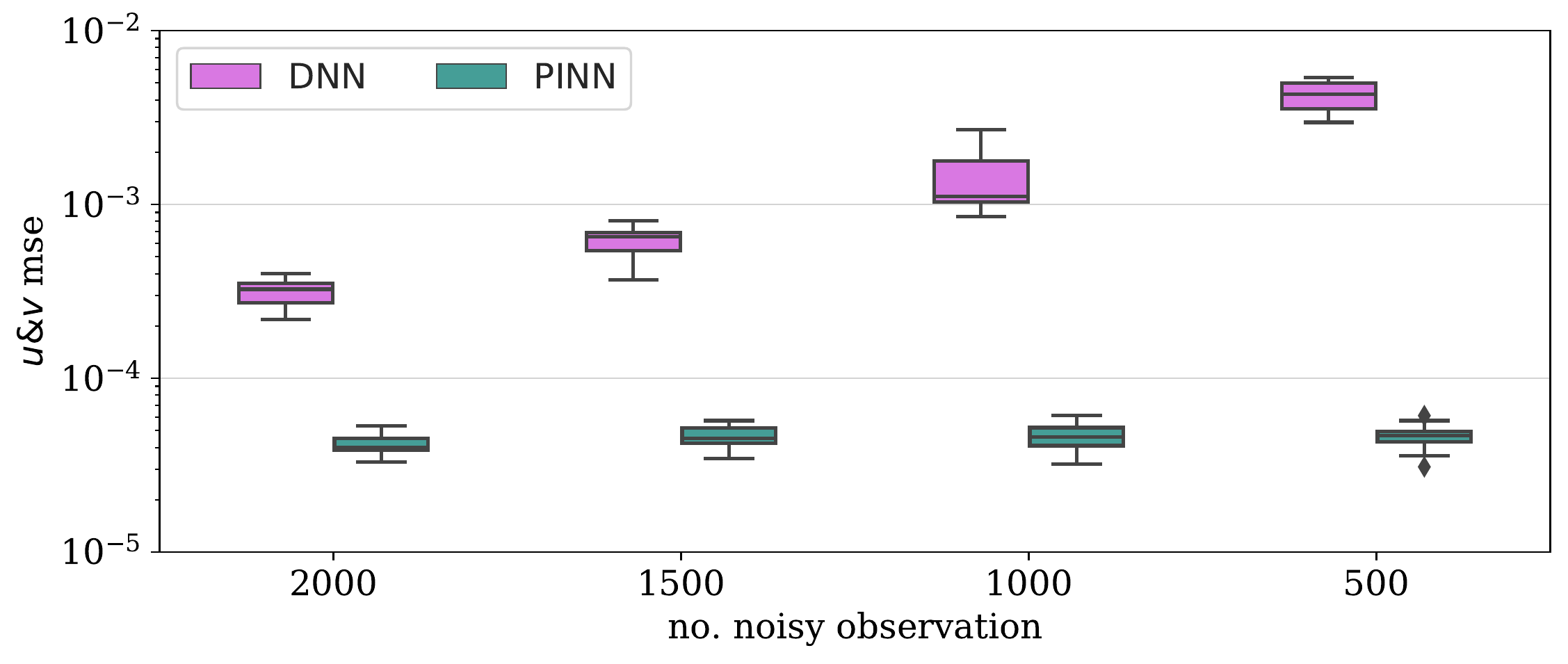}}
\caption{Boxplots of velocity test MSE for the scenarios where 500, 1000, 1500, and 2000 data points with 10\% noise are provided for model training.}
\label{fig:nobs-uv-mse}
\end{center}
\end{figure}

These data-sets of different sizes are similarly provided to both a DNN and a PINN for training. Test MSEs are computed for all the scenarios, and presented in Fig.~\ref{fig:nobs-uv-mse}.

\begin{figure*}[htbp]
\begin{center}
\centerline{\includegraphics[width=0.9\linewidth]{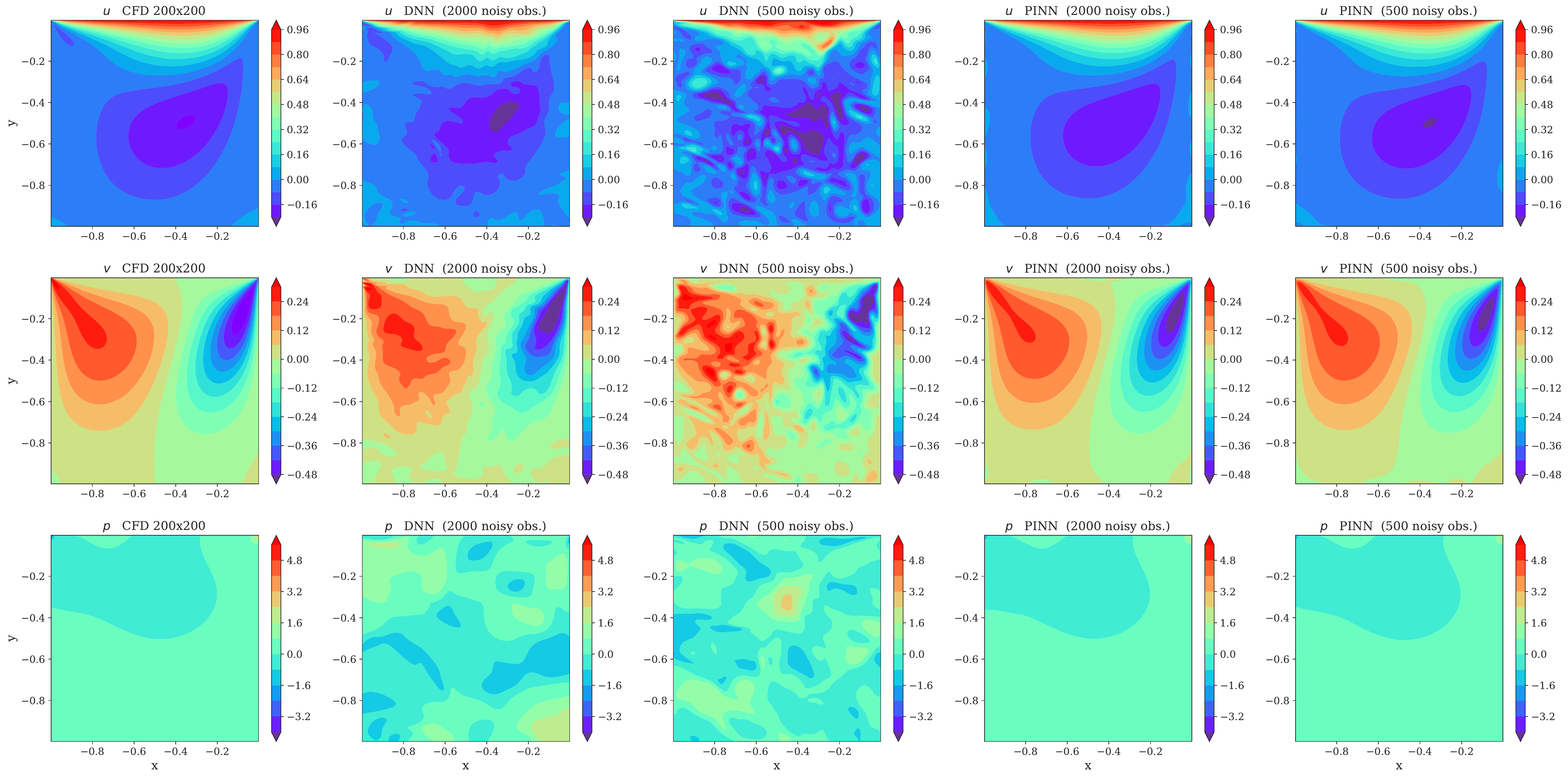}}
\caption{Representative contour plots of the $u$-velocity (Row 1), $v$-velocity (Row 2), and pressure (Row 3) for the (L-R) ground truth, DNN with 2000 and 500 data points, and PINN with 2000 and 500 data points. Data points have 10\% noise present.}
\label{fig:nobs-contour-pinns}
\end{center}
\end{figure*}

Representative contour plots of the velocity and pressure predicted by the DNN and PINN models are presented in Fig.~\ref{fig:nobs-contour-pinns} to illustrate the effectiveness of the PINN models relative to the DNN models for datasets of varying sizes.

The results show that the introduction of physics into the neural network is increasingly helpful as the data gets more noisy and sparse, with at least an order of magnitude improvement in the test error between the DNN and PINN. In this instance, we note orders of magnitude reduction in model test error across all cases. Importantly, even for this relatively simpler case study, we note that a halving of the amount of data provided leads to an approximately order of magnitude increase in the DNN test error. It's worth noting that the reduction in dataset size by 5x from 2500 to 500 degrades the DNN model performance to that equivalent to the use of low-quality sensors with noise larger than 50\%. Such trade-offs between the number and quality of sensors that can be acquired and deployed are dependent on the complexity of the problem, but are nonetheless important considerations in real-world problems.

Critically, the use of a PINN can overcome this reduction in the amount of data provided and prevent any significant performance degradation. This is thus extremely promising for the use of PINNs with scarce data-sets, especially in situations where the cost of sensors or data acquisition can severely reduce the availability of data. 

\section{Conclusion}

In this study, we report that the use of physics-based regularization is surprisingly effective for ameliorating noise in the data acquired. In particular, the final prediction error from trained models is reduced by orders of magnitude via the incorporation of underlying governing physical laws into the training process.

More specifically, the inclusion of physics-based regularization effectively overcomes noise that can be range up to equivalent SNR of 1. In this worst case scenario, the PINN models in this study still produce solutions that are approximately equivalent to models trained with data acquired by hypothetical sensors with 10x better noise performance. This effectiveness can be particularly significant for real-world applications, especially since 10x better sensors are frequently not just 10x more expensive, e.g. in chemical sensing.

Interestingly, we note that the PINN model as tested in this case study displays differential ability to overcome data scarcity, and data noise. Critically, the choice of sensors for real-world deployment and measurement frequently involves a multi-criteria optimization, whereby one wants to optimize both data quantity and quality under a limited budget (constraint). Understanding the relative sensitivity of PINN models to dataset size and noise can greatly impact the optimal sensor choice. Further tests on different scenarios and case studies to confirm the potential greater effectiveness of PINNs for resolving data noise can be beneficial for guiding this optimization. 

\section*{Acknowledgment}

This research is supported by A*STAR under the AME Programmatic programme$:$ "Explainable Physics-based AI for Engineering Modelling and Design (ePAI)" Award No. A20H5b0142 and the AI3 HTPO seed grant C211118016 on "Upside-Down Multi-Objective Bayesian Optimization for Few-Shot Design".





\bibliography{mybibfile}

\begin{thebibliography}{10}
\expandafter\ifx\csname url\endcsname\relax
  \def\url#1{\texttt{#1}}\fi
\expandafter\ifx\csname urlprefix\endcsname\relax\def\urlprefix{URL }\fi
\expandafter\ifx\csname href\endcsname\relax
  \def\href#1#2{#2} \def\path#1{#1}\fi

\bibitem{stewart2017label}
R.~Stewart, S.~Ermon, Label-free supervision of neural networks with physics
  and domain knowledge, in: Thirty-First AAAI Conference on Artificial
  Intelligence, 2017.

\bibitem{zhu2019physics}
Y.~Zhu, N.~Zabaras, P.-S. Koutsourelakis, P.~Perdikaris, Physics-constrained
  deep learning for high-dimensional surrogate modeling and uncertainty
  quantification without labeled data, Journal of Computational Physics 394
  (2019) 56--81.

\bibitem{karpatne2017theory}
A.~Karpatne, G.~Atluri, J.~H. Faghmous, M.~Steinbach, A.~Banerjee, A.~Ganguly,
  S.~Shekhar, N.~Samatova, V.~Kumar, Theory-guided data science: A new paradigm
  for scientific discovery from data, IEEE Transactions on Knowledge and Data
  Engineering 29~(10) (2017) 2318--2331.

\bibitem{von2019informed}
L.~Von~Rueden, S.~Mayer, J.~Garcke, C.~Bauckhage, J.~Schuecker, Informed
  machine learning--towards a taxonomy of explicit integration of knowledge
  into machine learning, Learning 18 (2019) 19--20.

\bibitem{le2021surrogate}
Q.~T. Le, C.~Ooi, Surrogate modeling of fluid dynamics with a multigrid
  inspired neural network architecture, Machine Learning with Applications 6
  (2021) 100176.

\bibitem{chen2020physics}
Y.~Chen, L.~Lu, G.~E. Karniadakis, L.~Dal~Negro, Physics-informed neural
  networks for inverse problems in nano-optics and metamaterials, Optics
  express 28~(8) (2020) 11618--11633.

\bibitem{raissi2020hidden}
M.~Raissi, A.~Yazdani, G.~E. Karniadakis, Hidden fluid mechanics: Learning
  velocity and pressure fields from flow visualizations, Science 367~(6481)
  (2020) 1026--1030.

\bibitem{yin2021non}
M.~Yin, X.~Zheng, J.~D. Humphrey, G.~E. Karniadakis, Non-invasive inference of
  thrombus material properties with physics-informed neural networks, Computer
  Methods in Applied Mechanics and Engineering 375 (2021) 113603.

\bibitem{chen2021physics}
Z.~Chen, Y.~Liu, H.~Sun, Physics-informed learning of governing equations from
  scarce data, Nature Communications 12~(1) (2021) 1--13.

\bibitem{chiu2022can}
P.-H. Chiu, J.~C. Wong, C.~Ooi, M.~H. Dao, Y.-S. Ong, {CAN-PINN}: A fast
  physics-informed neural network based on coupled-automatic--numerical
  differentiation method, Computer Methods in Applied Mechanics and Engineering
  395 (2022) 114909.

\bibitem{sun2020surrogate}
L.~Sun, H.~Gao, S.~Pan, J.-X. Wang, Surrogate modeling for fluid flows based on
  physics-constrained deep learning without simulation data, Computer Methods
  in Applied Mechanics and Engineering 361 (2020) 112732.

\bibitem{nabian2020physics}
M.~A. Nabian, H.~Meidani, Physics-driven regularization of deep neural networks
  for enhanced engineering design and analysis, Journal of Computing and
  Information Science in Engineering 20~(1) (2020) 011006.

\bibitem{yang2021b}
L.~Yang, X.~Meng, G.~E. Karniadakis, B-{PINN}s: Bayesian physics-informed
  neural networks for forward and inverse pde problems with noisy data, Journal
  of Computational Physics 425 (2021) 109913.

\bibitem{ghia1982high}
U.~Ghia, K.~N. Ghia, C.~Shin, High-{R}e solutions for incompressible flow using
  the {N}avier-{S}tokes equations and a multigrid method, Journal of
  computational physics 48~(3) (1982) 387--411.

\bibitem{jin2021nsfnets}
X.~Jin, S.~Cai, H.~Li, G.~E. Karniadakis, {NSFnets} ({N}avier-{S}tokes flow
  nets): {P}hysics-informed neural networks for the incompressible
  {N}avier-{S}tokes equations, Journal of Computational Physics 426 (2021)
  109951.

\bibitem{chiu2018improved}
P.-H. Chiu, An improved divergence-free-condition compensated method for
  solving incompressible flows on collocated grids, Computers \& Fluids 162
  (2018) 39--54.

\bibitem{chiu2021development}
P.-H. Chiu, H.~J. Poh, Development of an improved divergence-free-condition
  compensated coupled framework to solve flow problems with time-varying
  geometries, International Journal for Numerical Methods in Fluids 93~(1)
  (2021) 44--70.

\bibitem{wong2021learning}
J.~C. Wong, C.~Ooi, A.~Gupta, Y.-S. Ong, Learning in sinusoidal spaces with
  physics-informed neural networks, arXiv preprint arXiv:2109.09338 (2021).

\bibitem{le2021u}
T.~Q. Le, P.-H. Chiu, C.~Ooi, U-{N}et-based surrogate model for evaluation of
  microfluidic channels, International Journal of Computational Methods (2021)
  2141018.

\bibitem{wang2022dense}
H.~Wang, Y.~Liu, S.~Wang, Dense velocity reconstruction from particle image
  velocimetry/particle tracking velocimetry using a physics-informed neural
  network, Physics of Fluids 34~(1) (2022) 017116.

\end{thebibliography}



\end{document}